%% file: main.tex
\PassOptionsToPackage{table}{xcolor}
\documentclass[10pt, logo, onecolumn, copyright]{nv}

\usepackage{graphicx}
\usepackage{xcolor}
\usepackage[utf8]{inputenc}
\usepackage[T1]{fontenc}
\usepackage{times}
\usepackage{amsmath}
\usepackage{amssymb}
\usepackage{mathtools}
\usepackage{booktabs}
\usepackage{multirow}
\usepackage{tabularx}
\usepackage{subcaption}
\usepackage{float}
\usepackage{placeins}
\usepackage{url}
\usepackage[nameinlink]{cleveref}
\usepackage[square,sort,comma,numbers]{natbib}
\usepackage{xspace}

\definecolor{nvidiagreen}{HTML}{76B900}
\definecolor{nvgreen}{HTML}{D3E9AD}
\definecolor{nvgreenlight}{HTML}{EAF5D8}

\crefname{section}{Sec.}{Secs.}
\crefname{equation}{Eq.}{Eqs.}
\crefname{figure}{Figure}{Figures}
\crefname{table}{Table}{Tables}
\Crefname{figure}{Figure}{Figures}
\Crefname{table}{Table}{Tables}

\newcommand{\ourmethod}{SoL-Refiner\xspace}
\newcommand{\refbench}{Refiner-Bench\xspace}

\title{\ourmethod: Speed-of-Light One-Step \mbox{Refinement} for High-Resolution Video}

\author{
\parbox{\linewidth}{
\centering
\vspace{-5pt}
{\fontsize{9.6pt}{18pt}\selectfont\textbf{Haozhe Liu\textsuperscript{*}, ~ Tian Ye\textsuperscript{*}, ~ Shuchen Xue\textsuperscript{*}, ~ Sensen Gao\textsuperscript{*}, ~ Yitong Li, ~ Junsong Chen, ~ Haopeng Li}}
\\
\vspace{0.4em}
{\fontsize{9.6pt}{18pt}\selectfont\textbf{Jincheng Yu, ~ Duomin Wang, ~ Ruihua Zhang, ~ Lei Zhu, ~Jia-Wang Bian, ~ Song Han, ~ Enze Xie}}
\\
\vspace{2.5mm}
{\normalsize NVIDIA}
\\
\vspace{0.2em}
{\fontsize{8.5pt}{10pt}\selectfont \textsuperscript{*} Equal contribution.}
\\
\vspace{5pt}
{\fontsize{9pt}{11pt}\selectfont
\href{https://github.com/NVlabs/Sana/tree/sol-engine/models/sol-refiner}{\textcolor{nvidiagreen}{\raisebox{-0.15em}{\includegraphics[height=1em]{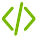}}\hspace{0.35em}\textbf{Code}}}
\hspace{1.8em}
\href{https://nvlabs.github.io/Sana/Sol-Refiner/}{\textcolor{nvidiagreen}{\raisebox{-0.15em}{\includegraphics[height=1em]{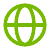}}\hspace{0.35em}\textbf{Project Page}}}
}
\\
\vspace{2pt}
}
}

\begin{abstract}
\vspace{-10pt}
High-resolution video generation is expensive, as its cost grows rapidly with the number of spatiotemporal tokens. A practical alternative first generates a lower-resolution video and then applies a refiner, but conventional multi-step refinement introduces a second sampling bottleneck. We present \textbf{\textcolor{nvidiagreen}{\ourmethod}}, a one-step video refiner that transforms low-resolution model outputs into 4K videos with a single denoising step. Our three-stage recipe combines high-resolution continual training, reinforcement learning (RL) post-training, and a final one-step distillation. We introduce \refbench, a video refinement benchmark constructed from the outputs of different video generators, and use a shared-input protocol to compare refiners at approximately 2K output resolution. With LTX-2.3 Refiner as the base model, the one-step \ourmethod outperforms all external refiners at 2K on the VBench and UniPercept averages, while at $3840\!\times\!2176$ it improves both metrics over the three-step LTX-2.3 Refiner. With the complete acceleration stack, \ourmethod achieves an $8.91\times$ speedup in refinement latency over the same baseline in our 2K latency setting. With MiniMax H3 as the base model, \ourmethod{} produces 2K videos in three refinement steps and is preferred over the official MiniMax H3 2K API in 87\% of user-study judgments.
\end{abstract}

\hypersetup{
  pdftitle={SoL-Refiner: Speed-of-Light One-Step Refinement for High-Resolution Video},
  pdfauthor={Haozhe Liu, Tian Ye, Shuchen Xue, Sensen Gao, Yitong Li, Junsong Chen, Haopeng Li, Jincheng Yu, Duomin Wang, Ruihua Zhang, Lei Zhu, Jia-Wang Bian, Song Han, Enze Xie}
}
\renewcommand{\today}{2026-11-04}
\begin{document}

\maketitle

\vspace{1pt}

\begin{figure}[h]
    \centering
    \begin{minipage}[t]{0.48\linewidth}
        \vspace{0pt}
        \begin{subfigure}[t]{\linewidth}
            \centering
            \includegraphics[width=\linewidth]{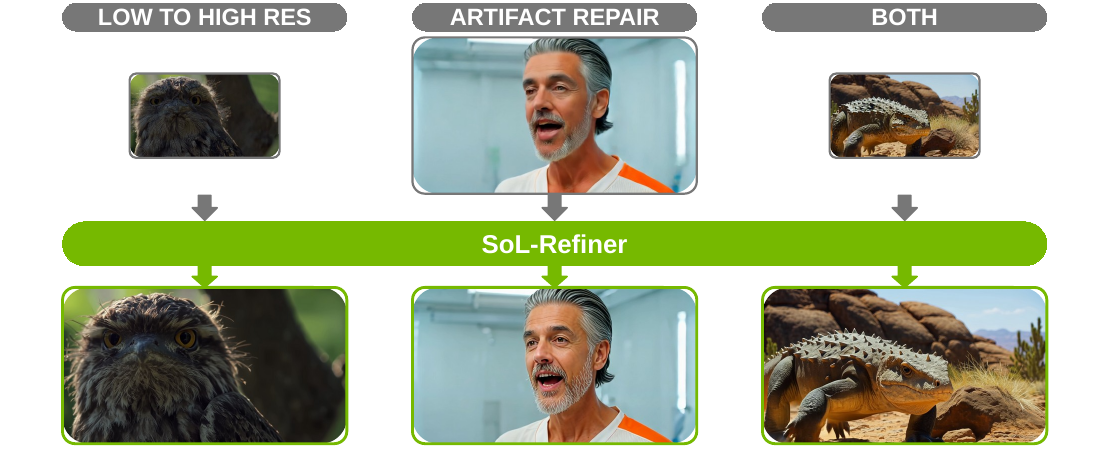}
            \caption{Resolution enhancement, artifact mitigation, and both.}
            \label{fig:overview-pipeline}
        \end{subfigure}
        \vspace{0mm}
        \begin{subfigure}[t]{\linewidth}
            \centering
            \includegraphics[width=\linewidth]{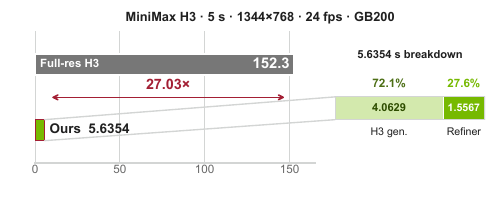}
            \caption{Full-resolution H3 versus distilled two-stage E2E latency.}
            \label{fig:overview-latency}
        \end{subfigure}
    \end{minipage}
    \hfill
    \begin{minipage}[t]{0.49\linewidth}
        \vspace{0pt}
        \begin{subfigure}[t]{\linewidth}
            \centering
            \includegraphics[width=\linewidth]{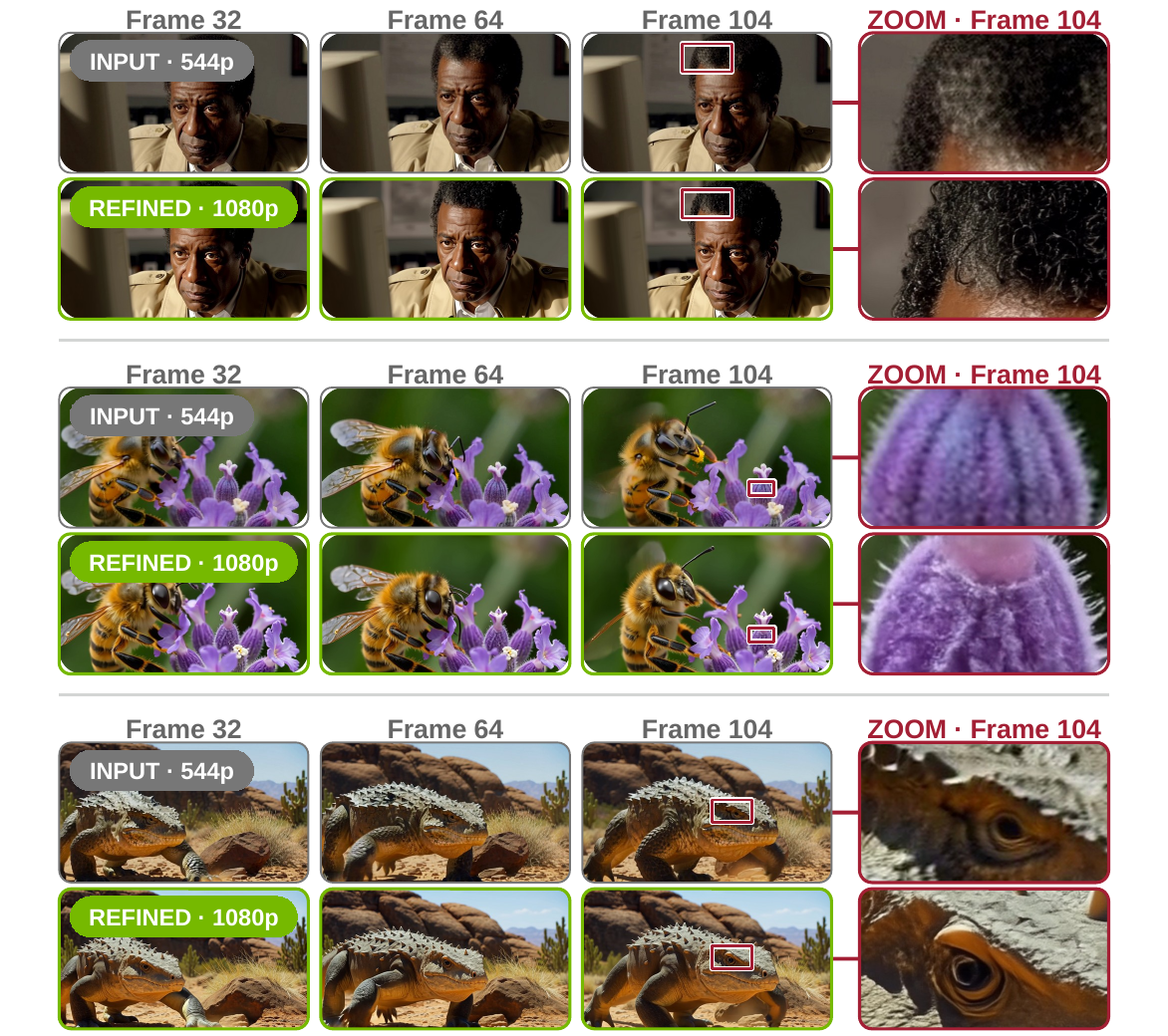}
            \caption{Qualitative visualization with magnified details.}
            \label{fig:overview-quality}
        \end{subfigure}
    \end{minipage}
    \vspace{-2mm}
    \caption{\textbf{Overview of \ourmethod.} (a) \ourmethod{} improves video resolution and visual quality while repairing artifacts in generated videos. (b) For a 5-second, $1344\!\times\!768$, 24-fps video, the published full-resolution 50-step SGLang MiniMax H3 baseline~\citep{li2026solh3} takes 152.3 s on one GB200, whereas distilled four-step $896\!\times\!512$ MiniMax H3 generation followed by the one-step \ourmethod{} takes 5.64 s using one GPU per stage ($27.03\times$). The two-stage pipeline spends 4.06 s in H3 generation and 1.56 s in Refiner service. (c) Qualitative examples show how \ourmethod{} improves low-resolution videos generated by MiniMax H3.}
    \label{fig:overview}
\end{figure}

\input{arxiv_sections/1_introduction}
\input{arxiv_sections/5_related_work}
\input{arxiv_sections/2_preliminary}
\input{arxiv_sections/3_method}
\input{arxiv_sections/4_experiments}
\FloatBarrier
\input{arxiv_sections/6_conclusion}

{
  \small
  \bibliographystyle{unsrtnat}
  \bibliography{venues_full,references}
}

\appendix
\input{arxiv_sections/6_appendix}

\end{document}

%% file: arxiv_sections/1_introduction.tex
\section{Introduction}
\label{sec:introduction}

Video generation has advanced toward higher visual fidelity and output resolution, but at substantial inference cost~\citep{ho2022imagenvideo,gao2025seedance,zhang2026flashvideo,kong2024hunyuanvideo}. Larger models increase the cost of each denoising step~\citep{kong2024hunyuanvideo,minimax2026h3}, while higher resolutions lengthen the spatiotemporal token sequence and raise the cost of every step.
Smaller models and lower-resolution sampling reduce this burden~\citep{chen2026sana,wan2025wanopenadvancedlargescale,ghafoorian2026mobilewan}, but can compromise fine texture and local detail~\citep{zhang2026flashvideo}.

A two-stage design addresses this trade-off by separating low-resolution content generation from high-resolution detail refinement~\citep{hacohen2026ltx2efficientjointaudiovisual,liu2024mardini,zhu2026sana}.
The base model establishes motion, composition, and scene content at low resolution, while the refiner enhances textures and local details at the target resolution.
The base stage can also use more aggressive acceleration, and the refiner repairs the resulting artifacts.
Existing refiners, however, often require multiple target-resolution denoising steps, creating a second sampling bottleneck~\citep{ma2026lingbot,hacohen2026ltx2efficientjointaudiovisual}.
Many refiners are also developed for a particular base generator, and their transfer to outputs from other generators is rarely evaluated.

We introduce \ourmethod{}, a one-step video refiner that transforms low-resolution inputs into high-resolution videos with improved local detail (\cref{fig:overview-pipeline,fig:overview-quality}).
It can refine outputs from different base generators without modifying or retraining the base models.
Starting from a pretrained LTX-2.3 checkpoint, training follows three stages: high-resolution continual training on paired videos, reinforcement learning (RL) post-training with frame-based reward models, and one-step distillation. We further reduce encoding, decoding, and denoising latency with a tiny autoencoder (TAE)~\citep{boerbohan2025taehv} and Sol-Engine~\citep{li2026solvideoinferenceengine}.
Within the refinement stage alone, the complete acceleration stack achieves an $8.91\times$ speedup over the three-step LTX-2.3 Refiner in our 2K latency setting. Paired with a four-step MiniMax H3 base model, the full two-stage pipeline is $27\times$ faster than direct full-resolution H3 generation (\cref{fig:overview-latency}).

Our contributions are threefold:
\begin{itemize}
    \item We introduce \ourmethod{}, a video refiner that upsamples and refines low-resolution outputs from multiple base generators in a single denoising step.
    \item We develop a three-stage training recipe that establishes high-resolution refinement through continual training, improves perceptual quality through RL post-training with frame-based rewards, and distills the resulting multi-step refiner into a one-step model.
    \item We introduce \refbench{}, a video refinement benchmark constructed from the outputs of different video generators, and use a shared-input protocol to compare refiners. At 2K resolution, \ourmethod{} outperforms the evaluated external refiners on average VBench and UniPercept scores, and improves over the three-step LTX-2.3 Refiner from 720p to 4K.
\end{itemize}

%% file: arxiv_sections/5_related_work.tex
\section{Related Work}
\label{sec:related-work}

\paragraph{Generated Video Refinement.}
\ourmethod{} targets synthesis artifacts in AI-generated videos, including distorted local structures and inconsistent textures. Resolution enlargement is optional: refinement can improve a generated video at its existing resolution or accompany upsampling (\cref{fig:overview-pipeline,fig:overview-quality}). Prior work also addresses generated video quality. VEnhancer combines spatial and temporal super-resolution with video enhancement~\citep{he2024venhancer}. Ultra Flash trains on degradations tailored to generated videos and combines reward-enhanced one-step distillation with cascaded streaming preference optimization~\citep{luxury2026ultraflash}. Its design centers on high-resolution streaming generation. Our focus is the refinement of generator outputs, including artifact correction without spatial enlargement, and we assess transfer across base generators.

\paragraph{Cascaded Generation.}
Cascaded generators use a low-resolution stage for content and motion, followed by high-resolution processing. Imagen Video interleaves spatial and temporal super-resolution models~\citep{ho2022imagenvideo}; FlashVideo learns a few-step flow-matching detail model~\citep{zhang2026flashvideo}; and LUVE combines latent upsampling with high-resolution experts~\citep{zhao2026luve}. The released LTX-2.3 and LingBot-Video pipelines use three-step and eight-step refiners, respectively~\citep{hacohen2026ltx2efficientjointaudiovisual,ma2026lingbot}. \ourmethod{} can serve as the refinement stage in such a cascade, using one denoising step to improve the base video. We evaluate released refiners on shared inputs from multiple generators to separate their refinement performance from the choice of upstream generator.

\paragraph{Video Restoration and SR.}
Diffusion-based restoration provides relevant priors and training methods for refinement. Upscale-A-Video uses recurrent latent propagation for temporally consistent upscaling~\citep{zhou2023upscaleavideo}, and SeedVR supports variable video lengths and resolutions through shifted-window attention~\citep{wang2025seedvr}. One-step methods reduce repeated denoising: DOVE uses staged latent-pixel training~\citep{chen2025dove}, UltraVSR uses a degradation-aware schedule and distillation~\citep{liu2025ultravsr}, and SeedVR2 uses adversarial post-training~\citep{wang2025seedvr2}. FlashVSR combines one-step distillation with causal sparse attention and a lightweight decoder~\citep{zhuang2025flashvsr}. We include SeedVR2 as a one-step restoration baseline in our evaluation. Our task concerns defects introduced during video synthesis; recovering spatial resolution is one application of the refiner, rather than its defining objective.

\paragraph{Distillation and Rewards.}
Our training builds on distribution matching and reward-based post-training. DMD2 combines teacher distribution matching with adversarial supervision~\citep{yin2024dmd2}, SF-V uses adversarial training for single-forward video generation~\citep{zhang2024sfv}, and ImageReward introduces Reward Feedback Learning~\citep{xu2023imagereward}. In Ultra Flash, perceptual rewards directly supervise the student during one-step distillation~\citep{luxury2026ultraflash}. We instead apply frame-based rewards to the multi-step refiner before distilling it through a few-step intermediate student. Distillation combines the teacher's distribution supervision with implicit distribution alignment~\citep{ge2025senseflow} and a projected DiT discriminator, drawing on projected adversarial supervision in PiD~\citep{lu2026pid}.

%% file: arxiv_sections/2_preliminary.tex
\section{Preliminaries}
\label{sec:preliminaries}

\paragraph{Latent Video Diffusion.}
An encoder maps an $F$-frame video $x\in\mathbb{R}^{F\times H\times W\times 3}$ to a clean latent $z_0$ with $L\approx FHW/(s_t s_h^2)$ spatiotemporal tokens, where $s_t$ and $s_h$ are temporal and spatial compression factors. At timestep $t$, the noisy latent is
\begin{equation}
    z_t = \alpha_t z_0 + \sigma_t \epsilon,
    \qquad \epsilon \sim \mathcal{N}(0, I),
\end{equation}
where $\alpha_t$ and $\sigma_t$ control the signal and noise levels. Sampling conditioned on text $c$ requires repeated denoising network function evaluations (NFEs), each processing the full latent sequence. Denoising cost grows with the NFE count, model size, and token count.

\paragraph{Two-Stage Generation.}
\label{sec:problem-formulation}
We generate a low-resolution video with a base generator $G_\phi$ and refine it with a high-capacity model $R_\theta$:
\begin{equation}
    x_{\mathrm{low}} = G_{\phi}(c), \qquad
    \hat{x}_{\mathrm{high}} = R_{\theta}(x_{\mathrm{low}}).
\end{equation}
Only $N_R$ refiner NFEs process target-resolution tokens. We target $N_R=1$, recovering fine detail while preserving the source content and motion.

%% file: arxiv_sections/3_method.tex
\section{Method}
\label{sec:method}

\begin{figure}[!htbp]
    \centering
    \includegraphics[width=\linewidth]{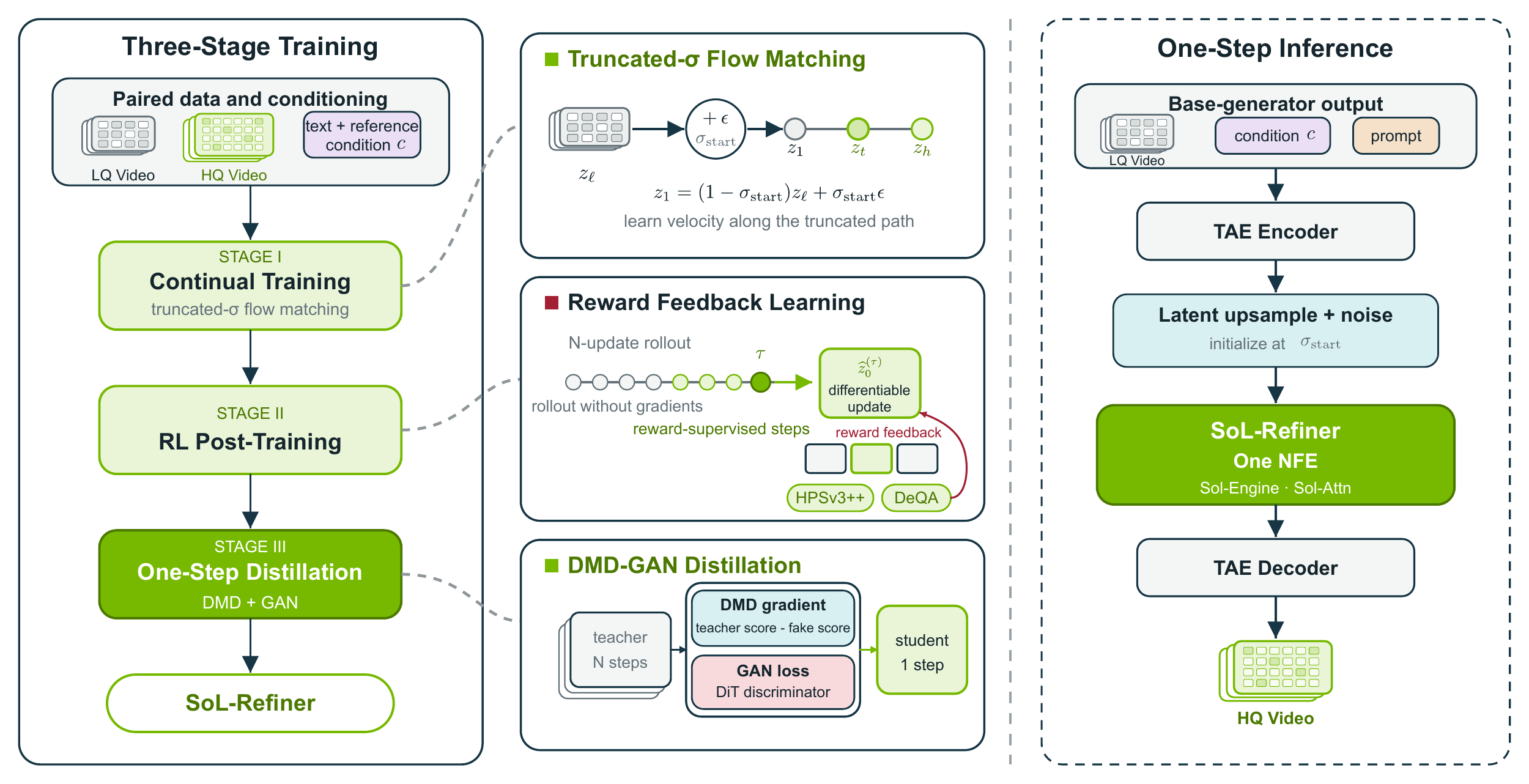}
    \caption{\textbf{Training and inference pipeline of \ourmethod.} Left, paired LQ/HQ videos and text/reference conditioning support continual training, RL post-training, and DMD-GAN distillation. Middle, the insets detail the truncated-$\sigma$ flow path, reward feedback after a no-gradient rollout, and the DMD gradient and DiT discriminator that compress a multi-step teacher into a one-step student. Right, a tiny autoencoder (TAE)~\citep{boerbohan2025taehv} replaces the full video variational autoencoder (VAE) to reduce encoding and decoding cost; latent upsampling and noise initialize the one-step refiner, and Sol Video Inference Engine (Sol-Engine)~\citep{li2026solvideoinferenceengine} executes one refiner NFE.}
    \label{fig:method-overview}
\end{figure}

\ourmethod{} maps a lower-resolution video $x_{\mathrm{low}}$ from a supported base generator to a target-resolution output, recovering local detail while preserving source content and motion. We train the refiner in three stages (\cref{fig:method-overview}): continual training learns the refinement mapping, RL post-training optimizes frame-based image-quality and preference signals, and distillation compresses the multi-step model into one denoising step.

\subsection{Stage I: Continual Training}

\paragraph{Paired Training Data.} Continual training uses paired conditioning and target videos to learn the refinement mapping. Each example contains a low-fidelity video $x_{\mathrm{cond}}$ and a high-fidelity target $x_{\mathrm{target}}$ with aligned content and motion. Our primary source of paired training data is an internal real-video dataset. We construct these pairs through data augmentation, retaining each real video as $x_{\mathrm{target}}$ and applying spatial downsampling followed by upsampling to obtain its low-fidelity counterpart $x_{\mathrm{cond}}$. We also use a small set of synthetic pairs only for initial warm-up. For these pairs, a supported base generator produces the low-resolution conditioning video $x_{\mathrm{cond}}$, which the LTX-2.3 Refiner refines into the corresponding high-fidelity target $x_{\mathrm{target}}$. The video encoder maps each pair to $z_{\ell}=\mathcal{E}(x_{\mathrm{cond}})$ and $z_h=\mathcal{E}(x_{\mathrm{target}})$.

\paragraph{Truncated-$\sigma$ Flow Matching.} Given $(z_{\ell},z_h)$, we construct a noisy source endpoint from the low-fidelity latent:
\begin{equation}
    z_1 = (1-\sigma_{\mathrm{start}})z_{\ell}
    + \sigma_{\mathrm{start}}\epsilon,
    \qquad \epsilon \sim \mathcal{N}(0,I),
    \label{eq:refiner-source}
\end{equation}
where $\sigma_{\mathrm{start}}=0.91$. We sample $\sigma_t$ from a shifted-logit-normal distribution truncated to $(0,\sigma_{\mathrm{start}}]$ and define
\begin{equation}
    \lambda_t = \frac{\sigma_t}{\sigma_{\mathrm{start}}},
    \qquad
    z_t = (1-\lambda_t)z_h + \lambda_t z_1.
    \label{eq:refiner-path}
\end{equation}
The resulting states lie on the segment from the clean target $z_h$ to the noisy source $z_1$. Parameterizing this path by $\sigma_t$ gives the target velocity
\begin{equation}
    v^{\star}
    = \frac{\partial z_t}{\partial \sigma_t}
    = \frac{z_1-z_h}{\sigma_{\mathrm{start}}}.
\end{equation}
We train the refiner with the flow-matching objective~\citep{lipman2022flow,liu2022flow}:
\begin{equation}
    \mathcal{L}_{\mathrm{ref}}
    = \mathbb{E}_{(z_{\ell},z_h,c),\,\sigma_t,\,\epsilon}
    \left[
        \left\|v_{\theta}(z_t,\sigma_t,c)-v^{\star}\right\|_2^2
    \right],
\end{equation}
where $c$ collects the conditioning signals. The truncated path retains information from $z_{\ell}$ at its noisiest endpoint, directing the learned vector field toward refinement rather than reconstruction from pure noise.

\paragraph{Conditioning and Initialization.} The conditioning set $c$ contains text and reference-image features. Reference-image tokens are concatenated with the video sequence but excluded from $\mathcal{L}_{\mathrm{ref}}$, allowing them to guide refinement without becoming reconstruction targets. We initialize the refiner from a pretrained high-capacity video diffusion model. The resulting multi-step refiner initializes Stage II.

\subsection{Stage II: RL Post-Training}

The Stage I flow-matching objective does not directly optimize perceptual quality or human preference. We therefore post-train the multi-step refiner using Reward Feedback Learning (ReFL)~\citep{xu2023imagereward}. Starting from a noisy low-fidelity latent, we first construct a truncated denoising schedule containing $N$ effective Euler updates. Since clean-latent predictions at early, high-noise states are insufficiently reliable for reward models, we restrict reward supervision to a set of later updates, $\mathcal{T}_{\mathrm{late}}=\{k,\ldots,N\}$, and uniformly sample $\tau \sim \mathcal{U}\!\left(\mathcal{T}_{\mathrm{late}}\right).$

The updates preceding $\tau$ are rolled out using the current policy without retaining gradients. At the selected state $(z_\tau,\sigma_\tau)$, the refiner performs a differentiable conditional forward pass and directly estimates the clean latent by taking an Euler step from $\sigma_\tau$ to zero:
\begin{equation}
    \widehat{z}_0^{(\tau)}
    =
    z_\tau
    -
    \sigma_\tau
    v_\theta(z_\tau,\sigma_\tau,c).
\end{equation}
Rewards are evaluated on frames decoded from $\widehat{z}_0^{(\tau)}$ and backpropagated only through the selected update. During training, we partition the decoded video into three equally sized temporal segments and uniformly sample one frame from each of the first, middle, and final thirds. Let $\mathcal{I}_1$, $\mathcal{I}_2$, and $\mathcal{I}_3$ denote these temporal segments. The reward frame set is
\begin{equation}
    \mathcal{F}
    =
    \{f_1,f_2,f_3\},
    \qquad
    f_j \sim \mathcal{U}(\mathcal{I}_j).
\end{equation}
We score the sampled frames using two complementary reward models. HPSv3++~\citep{liu2026hpsv3++} measures prompt-conditioned human preference and perceptual quality, while DeQA~\citep{you2025teaching} evaluates degradation-aware visual quality.
\begin{equation}
    \mathcal{R}
    =
    \frac{1}{3|\mathcal{F}|}
    \sum_{i\in\mathcal{F}}\min(h_i,10)
    +
    \frac{2}{3|\mathcal{F}|}
    \sum_{i\in\mathcal{F}}\min(d_i,4.5),
\end{equation}
where the coefficients correspond to normalized HPSv3++ and DeQA weights of $0.2$ and $0.4$, respectively. The upper clipping limits the influence of outlier scores and reduces reward exploitation.

In our final training configuration, the truncated training schedule contains $N=12$ effective updates and we sample $\tau$ uniformly from $\{6,\ldots,12\}$. Evaluation uses the full 23-update schedule.

\subsection{Stage III: Distillation and Acceleration}

Stage III progressively compresses the Stage II refiner. A few-step stage first learns a three-step student and its fake-score model. The one-step stage then initializes from this student and adds adversarial supervision while retaining the distribution-matching objective.

\paragraph{Few-Step Distillation.} We use distribution matching distillation (DMD)~\citep{yin2024dmd2} with three model roles: a trainable student generator $G_\theta$, a frozen real-score teacher $T$, and a trainable fake-score model $F_\phi$. We initialize all three models from the Stage II checkpoint and freeze $T$. We reuse the noisy source $z_1$ and interpolation path in \cref{eq:refiner-source,eq:refiner-path}, with $\sigma_{\mathrm{start}}=0.91$ for few-step distillation. We sample the generator noise level $\sigma_g$ from $\{0.91,0.73,0.42\}$ and denote the corresponding interpolated state by $z_g$. The student predicts the clean target latent:
\begin{equation}
    \widehat{z}_h=G_\theta(z_g,\sigma_g,c).
\end{equation}
For score evaluation, we replace the clean endpoint $z_h$ in \cref{eq:refiner-path} with $\widehat{z}_h$ and sample $\sigma_s\in[0.42,\sigma_{\mathrm{start}}]$. We query $T$ and $F_\phi$ on the resulting shared state $z_s$. Let $\widehat{z}_{\mathrm{real}}$ and $\widehat{z}_{\mathrm{fake}}$ denote their clean-latent predictions. We normalize their difference and apply it through a surrogate regression loss:
\begin{equation}
    g_{\mathrm{DMD}}
    =
    \frac{\widehat{z}_{\mathrm{fake}}-\widehat{z}_{\mathrm{real}}}
    {\operatorname{mean}\left(\left|\widehat{z}_h-\widehat{z}_{\mathrm{real}}\right|\right)+\varepsilon_g},
    \qquad
    \mathcal{L}_{\mathrm{DMD}}
    =
    \frac{1}{2}
    \left\|
    \widehat{z}_h-\operatorname{sg}\left(\widehat{z}_h-g_{\mathrm{DMD}}\right)
    \right\|_2^2,
\end{equation}
where $\operatorname{sg}$ stops gradients and $\varepsilon_g>0$ stabilizes the normalization. The fake-score model learns the student distribution from detached generator samples using the native flow-matching target, while $G_\theta$ receives an update every five fake-score updates. After each generator update, implicit distribution alignment (IDA)~\citep{ge2025senseflow} moves the fake-score parameters toward the generator,
\begin{equation}
    \phi \leftarrow \beta\phi+(1-\beta)\theta,
    \qquad \beta=0.97.
\end{equation}
We maintain an exponential moving average (EMA) of $G_\theta$ and use the resulting three-step checkpoint to initialize one-step distillation.

\paragraph{One-Step Distillation.} We initialize the one-step generator from the three-step student and initialize both score models from the Stage II teacher. We set $\sigma_{\mathrm{start}}=0.73$ and fix $\sigma_g=\sigma_{\mathrm{start}}$, matching the one-step inference schedule $[0.73,0]$. For score evaluation, we sample $\sigma_s\sim\mathcal{U}(\sigma_{\min},\sigma_{\max})$, where $0<\sigma_{\min}<\sigma_{\max}<\sigma_{\mathrm{start}}$. Fake-score training samples noise levels from the same uniform distribution. The generator input is therefore the noisy source $z_1$ from \cref{eq:refiner-source}, and each training sample requires one generator evaluation:
\begin{equation}
    \widehat{z}_h^{(1)}
    =
    G_{\theta_1}\left(z_1,\sigma_{\mathrm{start}},c\right).
\end{equation}
The one-step generator retains $\mathcal{L}_{\mathrm{DMD}}$ and adds a projected DiT discriminator for direct supervision from high-quality latents~\citep{yin2024dmd2,lu2026pid}. We reuse the frozen real-score DiT as the feature extractor and attach a separate lightweight discriminator. It pools token features from blocks $\{21,34,47\}$ with noise- and text-conditioned query heads, then maps the pooled features to one real-or-fake logit. This separation prevents the adversarial objective from changing the fake-score model used by the DMD gradient.

For each adversarial update, we sample a noise level $\sigma_a$ and construct real and fake states along the same low-quality-conditioned bridge. The states share the low-quality endpoint, Gaussian noise, and $\sigma_a$, while their clean endpoints are $z_h$ and $\widehat{z}_h^{(1)}$. We combine $\mathcal{L}_{\mathrm{DMD}}$ with a non-saturating generator loss and train the discriminator with the corresponding logistic real/fake loss.

We stabilize the discriminator with approximate R1 (aR1) regularization~\citep{lin2025apt}:
\begin{equation}
    \mathcal{L}_{\mathrm{aR1}}
    =
    \left\|D(z_a)-D(z_a+\delta)\right\|_2^2,
\end{equation}
where $z_a$ is the real bridge state and $\delta$ is a small Gaussian perturbation. Without regularization, the discriminator can separate real and fake bridge states too quickly, giving the one-step generator an unstable adversarial signal. Exact R1 requires second-order gradients through the DiT feature extractor, so aR1 approximates local smoothness by matching logits on a real state and its perturbation. We freeze the DiT feature extractor and alternate discriminator, fake-score, and generator updates, applying IDA after each generator update.

\subsection{Inference Optimization}

We accelerate refinement with Sol-Engine~\citep{li2026solvideoinferenceengine}, which combines kernel fusion with sparse attention based on Sol-Attn~\citep{li2026solattn} to reduce denoising computation. A tiny autoencoder (TAE)~\citep{boerbohan2025taehv} replaces the full VAE to reduce video encoding and decoding latency. For multi-step refinement, Sol-Engine additionally reuses cached intermediate results across denoising steps. \Cref{fig:latency-scaling} reports the latency gains for the multi-step refiner and the one-step pipeline.

%% file: arxiv_sections/4_experiments.tex
\section{Experiments}
\label{sec:experiments}

We evaluate \ourmethod{} on \refbench, which contains 150 videos. \Cref{sec:main-results} compares it with existing models across metrics and resolutions and evaluates acceleration across base generators. The H3 comparison in \Cref{fig:overview-latency} compares normal full-resolution MiniMax H3 generation with distilled low-resolution H3 followed by one-step refinement with \ourmethod{}. \Cref{sec:ablation-studies} analyzes the effect of each training stage and the latency gain from each component. More details about \refbench are in Appendix~\ref{app:refbench-composition}. We further extend \ourmethod{} to MiniMax H3~\citep{minimax2026h3} as an alternative base model; details are in Appendix~\ref{app:h3-extension}.

\subsection{Main Results}
\label{sec:main-results}

We compare \ourmethod{} with LingBot Stage-2 Refiner~\citep{ma2026lingbot}, LTX-2.3 Refiner~\citep{hacohen2026ltx2efficientjointaudiovisual}, LTX-2.0 Refiner~\citep{hacohen2026ltx2efficientjointaudiovisual}, and SEEDVR2~\citep{wang2025seedvr2} on \refbench. All 150 aligned videos are resized to $1024\!\times\!576$. Outputs are $1920\!\times\!1088$ for LingBot and $2048\!\times\!1152$ for the other refiners. VBench~\citep{huang2024vbench} averages subject consistency (SC), background consistency (BC), motion smoothness (MS), dynamic degree (DD), aesthetic quality (AQ), and imaging quality (IQ), while UniPercept~\citep{cao2025unipercept} averages IAA, IQA, and ISTA. With one refinement step, \ourmethod{} outperforms all evaluated external refiners on VBench AQ, IQ, and AVG and UniPercept AVG, including SEEDVR2 under the same step budget (\cref{tab:main_results,tab:unipercept_results}). Its VBench and UniPercept averages are 0.81048 and 60.4150, respectively. Our 23-step variant achieves higher scores on these metrics.

\begin{figure}[!t]
    \centering
    \includegraphics[width=\linewidth]{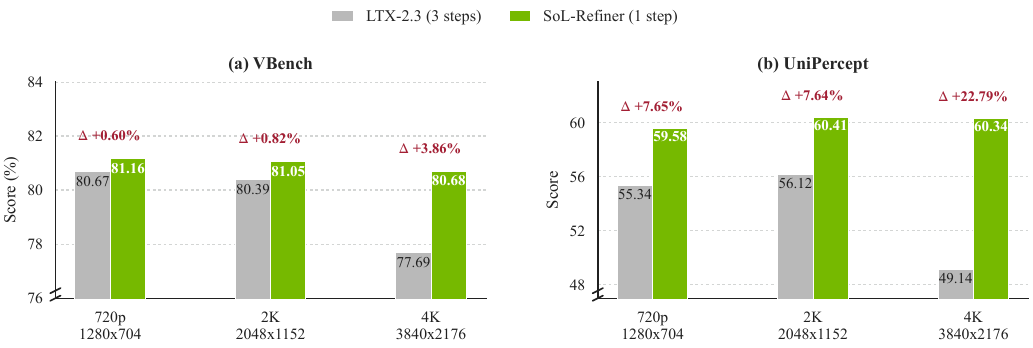}
    \caption{\textbf{Performance on \refbench across output resolutions.} We compare the three-step LTX-2.3 Refiner with the one-step \ourmethod{} using VBench AVG (left) and UniPercept AVG (right). Red $\Delta$ labels report the relative improvement over LTX-2.3 Refiner. VBench scores are shown on a 0--100 scale. Both vertical axes use truncated ranges for readability. Higher is better.}
    \label{fig:resolution_performance}
\end{figure}

We initialize \ourmethod{} from LTX-2.3 and therefore compare it with the official three-step LTX-2.3 Refiner as the direct baseline. \Cref{fig:resolution_performance} shows higher VBench and UniPercept scores at 720p, 2K, and 4K. At 4K, \ourmethod{} improves VBench AVG and UniPercept AVG over this baseline by 3.86\% and 22.79\%, respectively. In the 2K latency setting, the one-step model with TAE and Sol-Engine achieves an $8.91\times$ speedup in refinement latency (panel~(b) of \cref{fig:latency-scaling}).

\begin{table}[!t]
    \centering
    \normalfont\fontsize{7.2pt}{8.7pt}\selectfont
    \setlength{\tabcolsep}{2pt}
    \renewcommand{\arraystretch}{1.24}
    \caption{\textbf{VBench results on \refbench under a shared-input protocol.} All methods receive the same $1024\!\times\!576$ resized inputs. Resolution denotes the evaluated video resolution. SC, BC, MS, DD, AQ, and IQ denote subject consistency, background consistency, motion smoothness, dynamic degree, aesthetic quality, and imaging quality. $\Delta$ is the AVG difference from the resized input. Best results are in \textbf{bold}; second-best results are \underline{underlined}.}
    \label{tab:main_results}
    \begin{tabularx}{\linewidth}{@{}lcc *{8}{>{\centering\arraybackslash}X}@{}}
        \toprule
        \multirow{2}{*}{\textbf{Method}}
        & \multirow{2}{*}{\textbf{Eval. Res.}}
        & \multirow{2}{*}{\textbf{\# Steps}}
        & \multicolumn{7}{c}{\textbf{VBench}$\uparrow$}
        & \multirow{2}{*}{\textbf{$\Delta$}$\uparrow$} \\
        \cmidrule(lr{0.25em}){4-10}
        & & & \textbf{SC} & \textbf{BC} & \textbf{MS} & \textbf{DD} & \textbf{AQ} & \textbf{IQ} & \textbf{AVG} & \\
        \midrule
        Resized Input & $1024\!\times\!576$ & -- & 0.91805 & 0.95261 & 0.98354 & 0.70000 & 0.60621 & 0.63158 & 0.79866 & 0.00000 \\
        LingBot Stage-2 Refiner & $1920\!\times\!1088$ & 8 & 0.91715 & \underline{0.95354} & 0.98398 & \textbf{0.74000} & 0.57762 & 0.63792 & 0.80170 & 0.00304 \\
        LTX-2.3 Refiner & $2048\!\times\!1152$ & 3 & \underline{0.93004} & \underline{0.95354} & \textbf{0.98873} & 0.70000 & 0.59541 & 0.65553 & 0.80388 & 0.00522 \\
        LTX-2.0 Refiner & $2048\!\times\!1152$ & 3 & \textbf{0.93036} & \textbf{0.95909} & \underline{0.98656} & \underline{0.72000} & 0.60679 & 0.64497 & 0.80796 & 0.00930 \\
        SEEDVR2 & $2048\!\times\!1152$ & 1 & 0.91804 & 0.94783 & 0.97723 & 0.71333 & 0.60129 & 0.66337 & 0.80351 & 0.00485 \\
        \midrule
        \rowcolor{nvgreenlight}
        \ourmethod{} (Multi-Step) & $2048\!\times\!1152$ & 23 & 0.92988 & 0.94935 & 0.98027 & \underline{0.72000} & \textbf{0.62344} & \textbf{0.69851} & \textbf{0.81691} & \textbf{0.01825} \\
        \rowcolor{nvgreen}
        \textbf{\ourmethod{} (One-Step)} & $2048\!\times\!1152$ & 1 & 0.92191 & 0.94591 & 0.98054 & 0.71333 & \underline{0.60921} & \underline{0.69196} & \underline{0.81048} & \underline{0.01182} \\
        \bottomrule
    \end{tabularx}
\end{table}

\begin{table}[!t]
    \centering
    \normalfont\scriptsize
    \setlength{\tabcolsep}{4pt}
    \renewcommand{\arraystretch}{1.24}
    \caption{\textbf{UniPercept results on \refbench under a shared-input protocol.} All methods receive the same $1024\!\times\!576$ resized inputs. Resolution denotes the evaluated video resolution. Best results are in \textbf{bold}; second-best results are \underline{underlined}.}
    \label{tab:unipercept_results}
    \begin{tabularx}{\linewidth}{@{}lcc *{4}{>{\centering\arraybackslash}X}@{}}
        \toprule
        \multirow{2}{*}{\textbf{Method}}
        & \multirow{2}{*}{\textbf{Eval. Res.}}
        & \multirow{2}{*}{\textbf{\# Steps}}
        & \multicolumn{4}{c}{\textbf{UniPercept}$\uparrow$} \\
        \cmidrule(l{0.25em}){4-7}
        & & & \textbf{IAA} & \textbf{IQA} & \textbf{ISTA} & \textbf{AVG} \\
        \midrule
        Resized Input & $1024\!\times\!576$ & -- & 59.3540 & 60.8943 & 42.0890 & 54.1124 \\
        LingBot Stage-2 Refiner & $1920\!\times\!1088$ & 8 & 58.9189 & 58.0827 & 42.1403 & 53.0473 \\
        LTX-2.3 Refiner & $2048\!\times\!1152$ & 3 & 61.2239 & 61.9438 & \textbf{45.2052} & 56.1243 \\
        LTX-2.0 Refiner & $2048\!\times\!1152$ & 3 & 60.5074 & 60.4902 & 42.3823 & 54.4600 \\
        SEEDVR2 & $2048\!\times\!1152$ & 1 & 62.5692 & 64.4827 & 43.4057 & 56.8192 \\
        \midrule
        \rowcolor{nvgreenlight}
        \ourmethod{} (Multi-Step) & $2048\!\times\!1152$ & 23 & \textbf{66.7284} & \textbf{72.3977} & 44.8249 & \textbf{61.3170} \\
        \rowcolor{nvgreen}
        \textbf{\ourmethod{} (One-Step)} & $2048\!\times\!1152$ & 1 & \underline{65.7322} & \underline{70.6682} & \underline{44.8445} & \underline{60.4150} \\
        \bottomrule
    \end{tabularx}
\end{table}

\begin{figure}[!t]
    \centering
    \includegraphics[width=\linewidth]{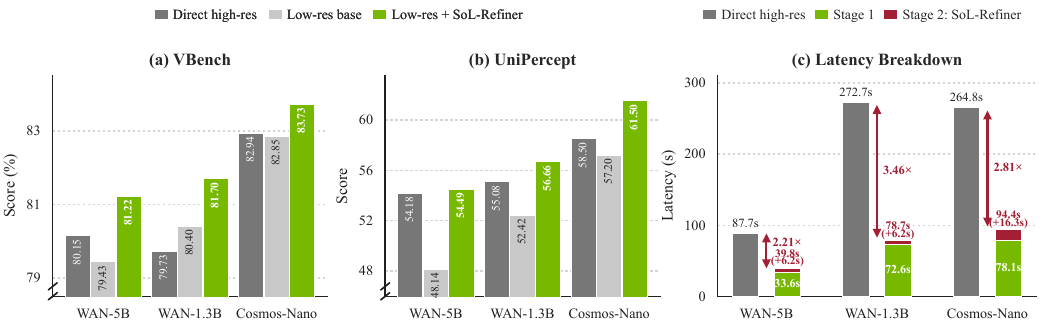}
    \caption{\textbf{Quality and acceleration across base generators.} (a,b) VBench and UniPercept averages for direct high-resolution generation, low-resolution generation, and low-resolution generation followed by SoL-Refiner. (c) End-to-end latency on one H100 GPU, split into base generation and refinement; speedups are relative to direct high-resolution generation. WAN uses 81 frames, 50 generation steps, and $832\!\times\!480\!\rightarrow\!1280\!\times\!704$ refinement; WAN-1.3B's direct baseline is $1280\!\times\!720$. Cosmos-Nano uses 189 frames, 35 generation steps, and 480p$\!\rightarrow\!1280\!\times\!720$ refinement.}
    \label{fig:base_generator_comparison}
\end{figure}

\paragraph{Acceleration across Base Generators.} We first run WAN-5B, WAN-1.3B, and Cosmos-Nano~\citep{agarwal2026cosmos} at low resolution, then use the one-step \ourmethod{} for upsampling and refinement. This moves multi-step generation to a smaller grid and uses only one denoising step at target resolution, thereby accelerating inference. As shown in \cref{fig:base_generator_comparison}, the pipeline reduces latency by 54.7\%, 71.1\%, and 64.4\%, respectively, while improving mean VBench and UniPercept over direct high-resolution generation. Exact values are reported in Appendix~\ref{app:base-generator-results}.

\subsection{Ablation Studies}
\label{sec:ablation-studies}

\paragraph{Training Recipe.} The training-recipe ablation evaluates continual training, RL post-training, and one-step DMD-GAN distillation. As shown in \cref{tab:training-recipe-ablation}, RL post-training gives the highest VBench and UniPercept averages. After distillation, the one-step model remains above the continual-training baseline on both averages.

\begin{table}[H]
    \centering
    \normalfont\fontsize{6.4pt}{7.8pt}\selectfont
    \setlength{\tabcolsep}{1.1pt}
    \renewcommand{\arraystretch}{1.20}
    \caption{\textbf{Ablation of the three-stage training recipe on \refbench.} We report VBench and UniPercept scores after each stage. Best results are in \textbf{bold}; second-best results are \underline{underlined}.}
    \label{tab:training-recipe-ablation}
    \begin{tabularx}{\linewidth}{@{}l *{11}{>{\centering\arraybackslash}X}@{}}
        \toprule
        \multirow{2}{*}{\textbf{Training Stage}}
        & \multicolumn{7}{c}{\textbf{VBench}$\uparrow$}
        & \multicolumn{4}{c}{\textbf{UniPercept}$\uparrow$} \\
        \cmidrule(lr{0.25em}){2-8}\cmidrule(l{0.25em}){9-12}
        & \textbf{SC} & \textbf{BC} & \textbf{MS} & \textbf{DD} & \textbf{AQ} & \textbf{IQ} & \textbf{AVG}
        & \textbf{IAA} & \textbf{IQA} & \textbf{ISTA} & \textbf{AVG} \\
        \midrule
        Stage 1: Continual (Multi-Step)
        & \underline{0.92756} & \textbf{0.95361} & \textbf{0.98201} & \textbf{0.72000} & 0.60526 & 0.66484 & 0.80888
        & 63.0807 & 64.5693 & 42.3359 & 56.6620 \\
        \rowcolor{nvgreenlight}
        Stage 2: Post-Training (Multi-Step)
        & \textbf{0.92988} & \underline{0.94935} & 0.98027 & \textbf{0.72000} & \textbf{0.62344} & \textbf{0.69851} & \textbf{0.81691}
        & \textbf{66.7284} & \textbf{72.3977} & \underline{44.8249} & \textbf{61.3170} \\
        \rowcolor{nvgreen}
        \textbf{Stage 3: DMD-GAN (One-Step)}
        & 0.92191 & 0.94591 & \underline{0.98054} & \underline{0.71333} & \underline{0.60921} & \underline{0.69196} & \underline{0.81048}
        & \underline{65.7322} & \underline{70.6682} & \textbf{44.8445} & \underline{60.4150} \\
        \bottomrule
    \end{tabularx}
\end{table}

\begin{figure}[!htbp]
    \centering
    \includegraphics[width=\linewidth]{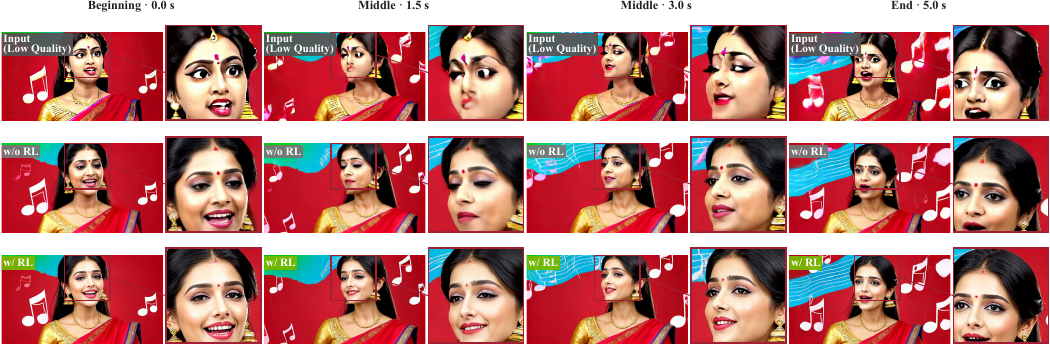}
    \caption{\textbf{Qualitative effect of RL post-training.} The rows show the low-quality input, the refiner without RL post-training, and the same refiner with RL post-training. Four aligned frames span 0.0 to 5.0 seconds. Each complete frame is followed by a magnified crop from the red box. In this selected example, RL post-training produces cleaner facial contours and sharper eyes, mouth, and jewelry across time.}
    \label{fig:rl-qualitative-main}
\end{figure}

\paragraph{Reward Learning.} We ablate the regularization recipe and the use of multiple reward models. \Cref{tab:reward-learning-ablation} shows that removing either design lowers the VBench average. In the selected video in \cref{fig:rl-qualitative-main}, RL post-training produces cleaner facial contours and details across four aligned frames. Appendix~\ref{app:rl-qualitative} provides two additional cases.

\begin{table}[H]
    \centering
    \normalfont\fontsize{6.9pt}{8.3pt}\selectfont
    \setlength{\tabcolsep}{1.8pt}
    \renewcommand{\arraystretch}{1.22}
    \caption{\textbf{Reward-learning ablation on \refbench.} We report VBench scores. The ``w/o multiple reward models'' variant uses only HPSv3++. Best results are in \textbf{bold}; second-best results are \underline{underlined}.}
    \label{tab:reward-learning-ablation}
    \begin{tabularx}{\linewidth}{@{}l *{7}{>{\centering\arraybackslash}X}@{}}
        \toprule
        \multirow{2}{*}{\textbf{Configuration}}
        & \multicolumn{7}{c}{\textbf{VBench}$\uparrow$} \\
        \cmidrule(l{0.25em}){2-8}
        & \textbf{SC} & \textbf{BC} & \textbf{MS} & \textbf{DD} & \textbf{AQ} & \textbf{IQ} & \textbf{AVG} \\
        \midrule
        w/o regularization recipe & \textbf{0.94438} & \textbf{0.95297} & \underline{0.98074} & 0.62667 & \textbf{0.64757} & \textbf{0.71044} & 0.81046 \\
        w/o multiple reward models & \underline{0.93553} & 0.94765 & \textbf{0.98443} & \underline{0.70000} & \underline{0.62616} & 0.69111 & \underline{0.81415} \\
        \midrule
        \rowcolor{nvgreen}
        \textbf{Ours} & 0.92988 & \underline{0.94935} & 0.98027 & \textbf{0.72000} & 0.62344 & \underline{0.69851} & \textbf{0.81691} \\
        \bottomrule
    \end{tabularx}
\end{table}

\paragraph{Distillation Ablation.} DMD-R jointly optimizes the RL and DMD objectives, whereas RL then DMD applies RL post-training before DMD. We evaluate both strategies in \cref{tab:distillation-ablation}. RL then DMD gives the higher VBench average. We therefore use RL then DMD in all other experiments.

\begin{table}[H]
    \centering
    \normalfont\fontsize{6.4pt}{7.8pt}\selectfont
    \setlength{\tabcolsep}{1.1pt}
    \renewcommand{\arraystretch}{1.20}
    \caption{\textbf{Distillation ablation on \refbench.} We compare DMD-R with RL followed by DMD using VBench. Best results are in \textbf{bold}; second-best results are \underline{underlined}.}
    \label{tab:distillation-ablation}
\begin{tabularx}{\linewidth}{@{}l *{7}{>{\centering\arraybackslash}X}@{}}
    \toprule
    \multirow{2}{*}{\textbf{Configuration}}
    & \multicolumn{7}{c}{\textbf{VBench}$\uparrow$} \\
    \cmidrule(lr{0.25em}){2-8}
    & \textbf{SC} & \textbf{BC} & \textbf{MS} & \textbf{DD}
    & \textbf{AQ} & \textbf{IQ} & \textbf{AVG} \\
    \midrule
        DMD-R
        & \underline{0.91385} & \underline{0.93886} & \underline{0.97779} & \underline{0.66000} & \textbf{0.61929} & \textbf{0.69608} & \underline{0.80098}
         \\
        \rowcolor{nvgreen}
        \textbf{RL then DMD}
        & \textbf{0.92191} & \textbf{0.94591} & \textbf{0.98054} & \textbf{0.71333} & \underline{0.60921} & \underline{0.69196} & \textbf{0.81048}
        \\
        \bottomrule
    \end{tabularx}
\end{table}

\begin{figure}[!htbp]
    \centering
    \includegraphics[width=\linewidth]{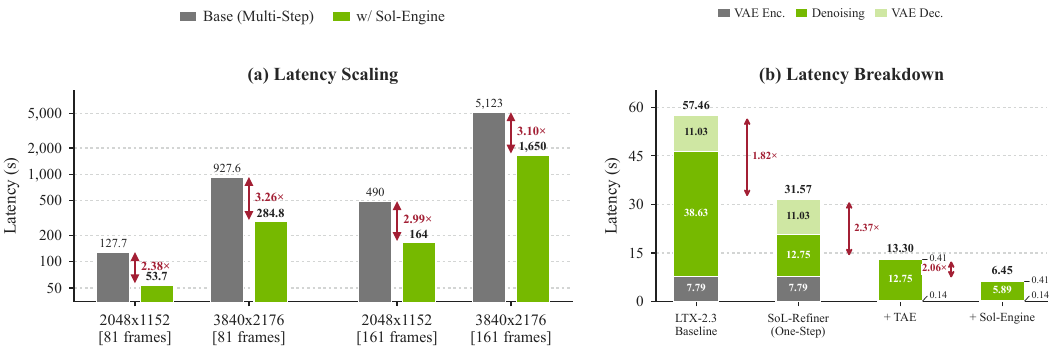}
    \caption{\textbf{Latency analysis.} Panel (a) compares Base (Multi-Step) with the same pipeline using Sol-Engine on one H100 GPU. The four groups report $2048\!\times\!1152$ and $3840\!\times\!2176$ outputs with 81 or 161 frames at 16 fps. Panel (b) reports component latency for $2048\!\times\!1152$ outputs with 241 frames at 24 fps. Each stacked bar shows VAE encoding, denoising, and VAE decoding. The first bar is the three-step LTX-2.3 Refiner baseline. The second bar is the one-step \ourmethod{}, and the final two bars add TAE and Sol-Engine. Red arrows report the speedup between adjacent configurations. The final configuration reduces refinement latency from 57.461 to 6.447 seconds, an $8.91\times$ speedup over the baseline. Lower latency is better.}
    \label{fig:latency-scaling}
\end{figure}

\paragraph{Latency Analysis.} We measure refinement latency on one H100 GPU and separate the contribution of each inference component. Panel (a) of \cref{fig:latency-scaling} compares the multi-step base model with Sol-Engine across four resolution and frame-count settings, where Sol-Engine gives $2.38\times$--$3.26\times$ speedups. Panel (b) starts from the three-step LTX-2.3 Refiner and adds one-step distillation, TAE, and Sol-Engine in sequence. These changes give adjacent speedups of $1.82\times$, $2.37\times$, and $2.06\times$ and reduce refinement latency from 57.461 to 6.447 seconds. The final configuration is $8.91\times$ faster than the baseline, showing that one-step distillation and system acceleration provide complementary gains. Appendix~\cref{tab:latency-scaling-details} reports the exact latency and VBench values.

\begin{figure}[!htbp]
    \centering
    \includegraphics[width=\linewidth]{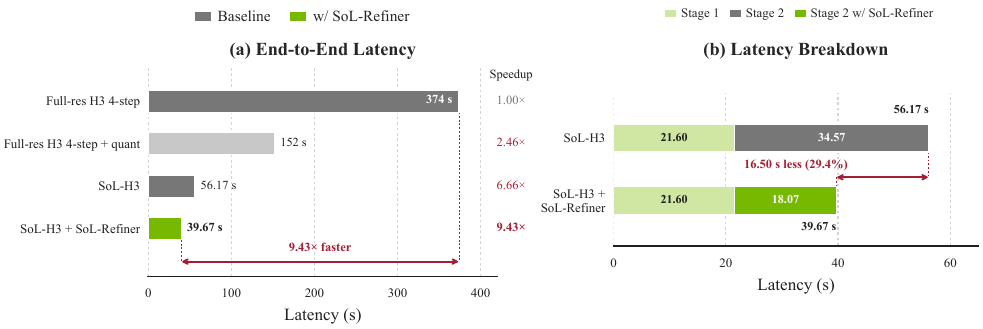}
    \caption{\textbf{H3 acceleration on a single NVIDIA DGX Spark (GB10).} (a) End-to-end latency for full-resolution four-step H3, its quantized variant, SoL-H3, and SoL-H3 with \ourmethod{}. Full-res denotes direct 768p generation; both SoL-H3 configurations generate at 384p and refine to 768p. Speedups use the 374-second full-resolution four-step baseline. (b) Stage-wise latency for the two SoL-H3 configurations. Stage~1 takes 21.60 seconds in both; Stage~2 decreases from 34.57 to 18.07 seconds with \ourmethod{}. The red annotation in panel (b) reports the 16.50-second reduction in total latency (29.4\% relative to SoL-H3). The panels use separate linear scales.}
    \label{fig:spark-latency}
\end{figure}

\paragraph{DGX Spark Deployment.} On a single NVIDIA DGX Spark (GB10), we evaluate SoL-H3~\citep{li2026solh3}, a two-stage pipeline with H3 generation at 384p (Stage~1) and refinement to 768p (Stage~2). Using \ourmethod{} in Stage~2, with Stage~1 unchanged, reduces end-to-end latency from 56.17 to 39.67 seconds (29.4\%), a $9.43\times$ speedup over direct four-step H3 generation at 768p (\cref{fig:spark-latency}).

%% file: arxiv_sections/6_conclusion.tex
\section{Conclusion}
\label{sec:conclusion}

\ourmethod{} makes high-resolution video generation more efficient by moving most denoising computation to a lower-resolution base generator and reserving one target-resolution step for refinement. Continual training learns the low-to-high-quality mapping, reward feedback learning improves perceptual quality, and staged DMD-GAN distillation with a projected DiT discriminator compresses refinement to one step. On \refbench{}, the one-step model outperforms external refiners in average VBench and UniPercept scores. In our 2K latency setting, the one-step refiner with TAE and Sol-Engine reduces refinement latency from 57.461 to 6.447 seconds, an $8.91\times$ speedup over the three-step LTX-2.3 Refiner.

%% file: arxiv_sections/6_appendix.tex
\clearpage
\section{Latency and Quality Comparison}
\label{app:latency-scaling-results}

\Cref{tab:latency-scaling-details} reports the exact latency and VBench values summarized in \cref{fig:latency-scaling}.

\begin{table}[!htbp]
    \centering
    \normalfont\fontsize{6.9pt}{8.3pt}\selectfont
    \setlength{\tabcolsep}{1.15pt}
    \renewcommand{\arraystretch}{1.20}
    \caption{\textbf{Latency and quality comparison.} Measurements use one H100 GPU. VBench metrics are evaluated on the corresponding 81-frame videos. Latency is reported in seconds per sample.}
    \label{tab:latency-scaling-details}
    \begin{tabularx}{\linewidth}{@{}c l c *{9}{>{\centering\arraybackslash}X}@{}}
        \toprule
        \multirow{2}{*}{\textbf{Resolution}}
        & \multirow{2}{*}{\textbf{Configuration}}
        & \multirow{2}{*}{\textbf{GPU}}
        & \multicolumn{7}{c}{\textbf{VBench}$\uparrow$}
        & \multicolumn{2}{c}{\textbf{Latency (s)}$\downarrow$} \\
        \cmidrule(lr{0.25em}){4-10}\cmidrule(l{0.25em}){11-12}
        & & & \textbf{SC} & \textbf{BC} & \textbf{MS} & \textbf{DD} & \textbf{AQ} & \textbf{IQ} & \textbf{AVG} & \shortstack{\textbf{81}\\\textbf{frames}} & \shortstack{\textbf{161}\\\textbf{frames}} \\
        \midrule
        & Base (Multi-Step) & H100 & 0.9082 & 0.9364 & 0.9813 & 0.7067 & 0.5618 & 0.6713 & 0.7943 & 127.7 & 490 \\
        \rowcolor{nvgreenlight}
        \multirow{-2}{*}{2K} & w/ Sol-Engine & H100 & 0.9092 & 0.9365 & 0.9813 & 0.7200 & 0.5550 & 0.6601 & 0.7937 & \textbf{53.7} & \textbf{164} \\
        \midrule
        & Base (Multi-Step) & H100 & 0.9037 & 0.9367 & 0.9806 & 0.7267 & 0.5547 & 0.6719 & 0.7957 & 927.6 & 5123 \\
        \rowcolor{nvgreenlight}
        \multirow{-2}{*}{4K} & w/ Sol-Engine & H100 & 0.9043 & 0.9386 & 0.9817 & 0.7267 & 0.5511 & 0.6591 & 0.7936 & \textbf{284.8} & \textbf{1650} \\
        \bottomrule
    \end{tabularx}
\end{table}

\section{\refbench Composition and Evaluation Details}
\label{app:refbench-composition}

\refbench contains 50 videos from each of WAN 2.1 1.3B~\citep{wan2025wanopenadvancedlargescale}, SANA-Video 2B~\citep{chen2026sana}, and LTX-2 Stage 1~\citep{hacohen2026ltx2efficientjointaudiovisual}. WAN and SANA-Video provide out-of-distribution inputs, while LTX-2 Stage 1 provides in-distribution inputs. The benchmark spans 12 content groups, three motion levels, and six camera-motion types. Each sample has one content-group, motion-level, and camera-motion label, and every content group includes samples from all three generators.

\paragraph{Source Preparation.} The native generator outputs are cropped to valid aligned sizes, then resized to a common $1024\!\times\!576$ input for the main 2K benchmark. \Cref{tab:refbench-source-preparation} records each stage. The resolution study instead uses a half-scale input for each target, including $1920\!\times\!1088$ for the 4K setting.

\begin{table}[!htbp]
    \centering
    \normalfont\scriptsize
    \setlength{\tabcolsep}{4pt}
    \renewcommand{\arraystretch}{1.14}
    \caption{\textbf{Source preparation for the main 2K \refbench setting.} Native and aligned resolutions are generator-specific, while every aligned video is resized to the shared $1024\!\times\!576$ refiner input.}
    \label{tab:refbench-source-preparation}
    \begin{tabularx}{\linewidth}{@{}l c *{3}{>{\centering\arraybackslash}X}@{}}
        \toprule
        \textbf{Source Generator} & \textbf{Videos} & \textbf{Native Res.} & \textbf{Aligned Crop} & \textbf{Main Refiner Input} \\
        \midrule
        WAN 2.1 1.3B & 50 & $832\!\times\!480$ & $832\!\times\!448$ & $1024\!\times\!576$ \\
        SANA-Video 2B & 50 & $1280\!\times\!736$ & $1280\!\times\!704$ & $1024\!\times\!576$ \\
        LTX-2 Stage 1 & 50 & $768\!\times\!512$ & $768\!\times\!512$ & $1024\!\times\!576$ \\
        \bottomrule
    \end{tabularx}
\end{table}

\paragraph{Compared Refiners.} We compare \ourmethod{} with LingBot Stage-2 Refiner~\citep{ma2026lingbot}, LTX-2.3 Refiner~\citep{hacohen2026ltx2efficientjointaudiovisual}, LTX-2.0 Refiner~\citep{hacohen2026ltx2efficientjointaudiovisual}, SEEDVR2~\citep{wang2025seedvr2}, and our multi-step model. This set covers one-step and multi-step refinement methods.

\paragraph{Evaluation Protocol and Metrics.} In the main comparison, all methods receive the same 150 aligned inputs resized to $1024\!\times\!576$. LingBot Stage-2 Refiner produces $1920\!\times\!1088$ outputs, while the other refiners produce $2048\!\times\!1152$ outputs. We report VBench~\citep{huang2024vbench} subject consistency, background consistency, motion smoothness, dynamic degree, aesthetic quality, imaging quality, and their mean. We also report UniPercept~\citep{cao2025unipercept} IAA, IQA, ISTA, and their mean. UniPercept uses eight uniformly sampled frames per video.

\paragraph{Content Groups.} \refbench covers people, animals, natural scenes, built environments, objects, and stylized content. \Cref{tab:refbench-content-groups} lists all groups and their main visual challenges.

\begin{table}[!htbp]
    \centering
    \normalfont\scriptsize
    \setlength{\tabcolsep}{5pt}
    \renewcommand{\arraystretch}{1.08}
    \caption{\textbf{Content groups in \refbench.} The descriptions summarize the main visual challenges represented by each group.}
    \label{tab:refbench-content-groups}
    \begin{tabularx}{\linewidth}{@{}l c X@{}}
        \toprule
        \textbf{Content Group} & \textbf{Videos} & \textbf{Main Visual Challenges} \\
        \midrule
        Human daily activity & 18 & Hands, object interaction, water, and reflections \\
        Animals & 15 & Fur, articulated motion, fast motion, and water \\
        Nature and landscape & 15 & Water, weather, foliage, and particles \\
        Sports and dance & 15 & Fast articulated motion, hands, and fine body geometry \\
        Cinematic portraits & 12 & Faces, skin, hair, hands, and low-light detail \\
        Urban and architecture & 12 & Structural lines, repeated geometry, and reflections \\
        Vehicles and machines & 12 & Rigid motion, mechanical parts, dust, and reflections \\
        Multi-subject interaction & 11 & Identity consistency, hand interaction, and occlusion \\
        Camera-motion stress & 10 & Viewpoint change, complex geometry, and reflections \\
        Fantasy, sci-fi, and stylized & 10 & Glow, transparency, unusual materials, and visual effects \\
        Food and cooking & 10 & Liquids, steam, smoke, and hands \\
        Product and object close-up & 10 & Fine geometry, reflections, and micro-motion \\
        \bottomrule
    \end{tabularx}
\end{table}

\paragraph{Motion Levels.} \refbench contains 50 low-motion, 64 medium-motion, and 36 high-motion videos. These labels provide a coarse summary of scene-motion complexity. Camera behavior is recorded separately.

\paragraph{Camera Motions.} The camera labels are dolly (34), orbit (30), pan (29), handheld (22), tilt (19), and static (16). Static clips use a fixed viewpoint. Pan and tilt rotate the view horizontally and vertically. Dolly translates the camera through the scene, while orbit moves the viewpoint around a subject. Handheld clips contain irregular local camera movement.

\section{Detailed Base-Generator Results}
\label{app:base-generator-results}

\Cref{tab:base-generator-details} reports the quality means and latency values summarized in \cref{fig:base_generator_comparison}. \Cref{tab:base-generator-vbench-details,tab:base-generator-unipercept-details} provide the complete metric breakdowns.

\begin{table}[!htbp]
    \centering
    \normalfont\scriptsize
    \setlength{\tabcolsep}{3.5pt}
    \renewcommand{\arraystretch}{1.18}
    \caption{\textbf{Detailed quality and latency results across base generators.} VBench is reported on a 0--100 scale. Latency and SoL-Refiner overhead are measured in seconds per sample on one H100 GPU. The overhead is the red Stage-2 segment in \cref{fig:base_generator_comparison}.}
    \label{tab:base-generator-details}
    \begin{tabularx}{\linewidth}{@{}llc *{4}{>{\centering\arraybackslash}X}@{}}
        \toprule
        \textbf{Generator} & \textbf{Configuration} & \textbf{Resolution} & \textbf{VBench}$\uparrow$ & \textbf{UniPercept}$\uparrow$ & \textbf{Latency}$\downarrow$ & \textbf{Overhead}$\downarrow$ \\
        \midrule
        \multirow{3}{*}{WAN-5B}
        & Direct high-res & $1280\!\times\!704$ & 80.15 & 54.18 & 87.70 & -- \\
        & Low-res base & $832\!\times\!480$ & 79.43 & 48.14 & 33.59 & -- \\
        \rowcolor{nvgreenlight}
        & Low-res + SoL-Refiner & $1280\!\times\!704$ & \textbf{81.22} & \textbf{54.49} & \textbf{39.75} & 6.16 \\
        \midrule
        \multirow{3}{*}{WAN-1.3B}
        & Direct high-res & $1280\!\times\!720$ & 79.73 & 55.08 & 272.69 & -- \\
        & Low-res base & $832\!\times\!480$ & 80.40 & 52.42 & 72.55 & -- \\
        \rowcolor{nvgreenlight}
        & Low-res + SoL-Refiner & $1280\!\times\!704$ & \textbf{81.70} & \textbf{56.66} & \textbf{78.71} & 6.16 \\
        \midrule
        \multirow{3}{*}{Cosmos-Nano}
        & Direct high-res & $1280\!\times\!720$ & 82.94 & 58.50 & 264.78 & -- \\
        & Low-res base & 480p & 82.85 & 57.20 & 78.07 & -- \\
        \rowcolor{nvgreenlight}
        & Low-res + SoL-Refiner & 720p & \textbf{83.73} & \textbf{61.50} & \textbf{94.38} & 16.31 \\
        \bottomrule
    \end{tabularx}
\end{table}

\begin{table}[!htbp]
    \centering
    \normalfont\scriptsize
    \setlength{\tabcolsep}{2.5pt}
    \renewcommand{\arraystretch}{1.16}
    \caption{\textbf{Detailed VBench results across base generators.} SC, BC, MS, DD, AQ, and IQ denote subject consistency, background consistency, motion smoothness, dynamic degree, aesthetic quality, and imaging quality. All metrics are reported on their original 0--1 scale.}
    \label{tab:base-generator-vbench-details}
    \begin{tabularx}{\linewidth}{@{}ll *{7}{>{\centering\arraybackslash}X}@{}}
        \toprule
        \textbf{Generator} & \textbf{Configuration} & \textbf{SC}$\uparrow$ & \textbf{BC}$\uparrow$ & \textbf{MS}$\uparrow$ & \textbf{DD}$\uparrow$ & \textbf{AQ}$\uparrow$ & \textbf{IQ}$\uparrow$ & \textbf{AVG}$\uparrow$ \\
        \midrule
        \multirow{3}{*}{WAN-5B}
        & Direct high-res & 0.92258 & 0.95235 & 0.98199 & 0.71333 & 0.58522 & 0.65348 & 0.80149 \\
        & Low-res base & 0.90244 & 0.92967 & 0.96584 & 0.82000 & 0.53374 & 0.61434 & 0.79434 \\
        \rowcolor{nvgreenlight}
        & Low-res + SoL-Refiner & 0.90250 & 0.94432 & 0.96426 & 0.82000 & 0.56545 & 0.67668 & \textbf{0.81220} \\
        \midrule
        \multirow{3}{*}{WAN-1.3B}
        & Direct high-res & 0.92994 & 0.95195 & 0.97886 & 0.71333 & 0.57893 & 0.63057 & 0.79727 \\
        & Low-res base & 0.92382 & 0.94247 & 0.97691 & 0.78000 & 0.57314 & 0.62767 & 0.80400 \\
        \rowcolor{nvgreenlight}
        & Low-res + SoL-Refiner & 0.92213 & 0.95378 & 0.97583 & 0.79333 & 0.59331 & 0.66346 & \textbf{0.81697} \\
        \midrule
        \multirow{3}{*}{Cosmos-Nano}
        & Direct high-res & 0.86547 & 0.92167 & 0.98619 & 0.92667 & 0.59915 & 0.67715 & 0.82938 \\
        & Low-res base & 0.87199 & 0.91853 & 0.98841 & 0.93333 & 0.59479 & 0.66395 & 0.82850 \\
        \rowcolor{nvgreenlight}
        & Low-res + SoL-Refiner & 0.87203 & 0.93118 & 0.98339 & 0.92000 & 0.60469 & 0.71251 & \textbf{0.83730} \\
        \bottomrule
    \end{tabularx}
\end{table}

\begin{table}[!htbp]
    \centering
    \normalfont\scriptsize
    \setlength{\tabcolsep}{4pt}
    \renewcommand{\arraystretch}{1.24}
    \caption{\textbf{Detailed UniPercept results across base generators.} IAA, IQA, and ISTA denote Image Aesthetics Assessment, Image Quality Assessment, and Image Structure and Texture Assessment, respectively.}
    \label{tab:base-generator-unipercept-details}
    \begin{tabularx}{\linewidth}{@{}ll *{4}{>{\centering\arraybackslash}X}@{}}
        \toprule
        \textbf{Generator} & \textbf{Configuration} & \textbf{IAA}$\uparrow$ & \textbf{IQA}$\uparrow$ & \textbf{ISTA}$\uparrow$ & \textbf{AVG}$\uparrow$ \\
        \midrule
        \multirow{3}{*}{WAN-5B}
        & Direct high-res & 60.6266 & 61.3029 & 40.5970 & 54.1755 \\
        & Low-res base & 52.4973 & 53.1629 & 38.7562 & 48.1388 \\
        \rowcolor{nvgreenlight}
        & Low-res + SoL-Refiner & 59.1759 & 64.2542 & 40.0342 & \textbf{54.4881} \\
        \midrule
        \multirow{3}{*}{WAN-1.3B}
        & Direct high-res & 62.5408 & 62.5327 & 40.1709 & 55.0815 \\
        & Low-res base & 59.1791 & 58.7869 & 39.2797 & 52.4152 \\
        \rowcolor{nvgreenlight}
        & Low-res + SoL-Refiner & 63.1862 & 66.1531 & 40.6515 & \textbf{56.6636} \\
        \midrule
        \multirow{3}{*}{Cosmos-Nano}
        & Direct high-res & 63.0820 & 64.5257 & 47.8925 & 58.5001 \\
        & Low-res base & 62.1208 & 63.1083 & 46.3756 & 57.2016 \\
        \rowcolor{nvgreenlight}
        & Low-res + SoL-Refiner & 65.5842 & 71.1243 & 47.8062 & \textbf{61.5049} \\
        \bottomrule
    \end{tabularx}
\end{table}

\section{Additional RL Post-Training Comparisons}
\label{app:rl-qualitative}

\begin{figure}[H]
    \centering
    \begin{subfigure}[t]{\linewidth}
        \centering
        \includegraphics[width=\linewidth]{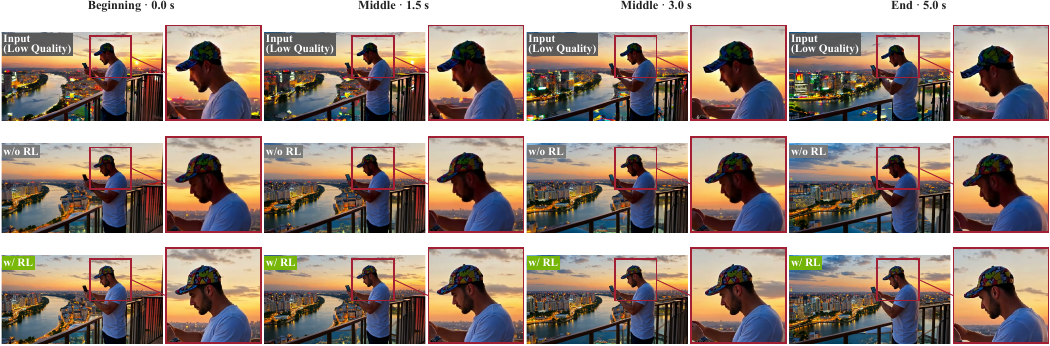}
        \caption{\textbf{Human subject.}}
        \label{fig:rl-qualitative-case09}
    \end{subfigure}
    \vspace{0.4em}
    \begin{subfigure}[t]{\linewidth}
        \centering
        \includegraphics[width=\linewidth]{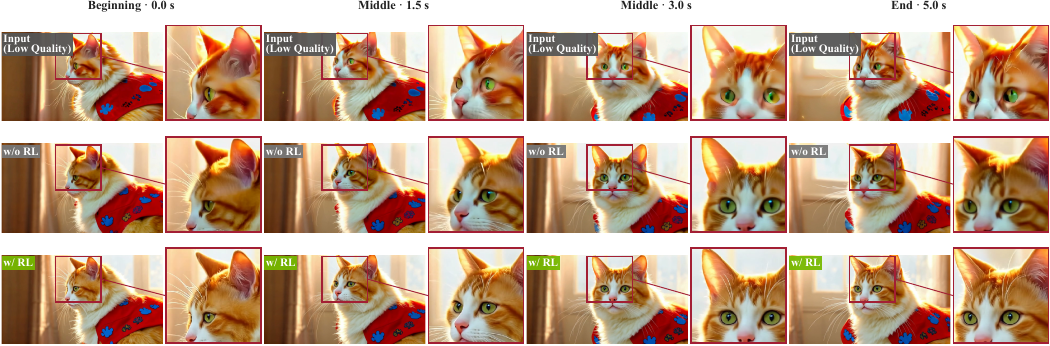}
        \caption{\textbf{Animal subject.}}
        \label{fig:rl-qualitative-case04}
    \end{subfigure}
    \caption{\textbf{Additional qualitative effects of RL post-training.} Panels (a) and (b) show selected human and animal examples. Each panel compares the low-quality input, the refiner without RL post-training, and the same refiner with RL post-training at four aligned frames from 0.0 to 5.0 seconds. Each complete frame is followed by a magnified crop from the red box. RL post-training sharpens the side profile and cap edges in panel (a), and improves the eyes, facial contours, and fur texture in panel (b).}
    \label{fig:rl-qualitative-additional}
\end{figure}

\Cref{fig:rl-qualitative-additional} provides two additional selected examples of the effect of RL post-training. The frame indices and crop coordinates are aligned across all three rows in both panels.

\enlargethispage{4\baselineskip}

\section{Extension to MiniMax H3 as the Base Model}
\label{app:h3-extension}

MiniMax H3~\citep{minimax2026h3} provides an official 2K regeneration API that refines its 768P outputs to $2560\!\times\!1440$. To show that our recipe transfers beyond LTX-2.3, we build \ourmethod{} on MiniMax H3 as the base model. The resulting refiner upgrades both MiniMax H3's own 768P drafts and low-resolution videos from other generators, such as WAN~2.1, SANA-Video, and LTX-2, to $2560\!\times\!1440$. We compare it with the official API on both kinds of inputs.

\paragraph{Setup.} We adopt MiniMax H3~\citep{minimax2026h3} as the base model and train only Stage~I of \ourmethod{}, a rank-384 LoRA on the frozen transformer. For few-step inference, we directly merge it with the community LightX2V Turbo acceleration LoRA~\citep{lightx2v2026minimaxh3turbo} and LMS Sharp LoRA~\citep{alissonerdx2026lmssharp}. The input video is encoded by the H3 VAE, upsampled in latent space by a community latent upscaler~\citep{lbh2026h3latentupscaler}, and refined to $2560\!\times\!1440$ in three steps. We use two kinds of inputs: the $1024\!\times\!576$ \refbench{} clips, with 50 clips from each of WAN~2.1, SANA-Video, and LTX-2, and 768P drafts generated by MiniMax H3 itself.

\paragraph{Quantitative Results.} We compare with the official MiniMax H3 2K API on \refbench{} in \cref{tab:h3-extension-results}. Our model improves the VBench average from 0.80242 to 0.81438 and the UniPercept average from 53.4317 to 59.9704, with higher per-clip averages on 125 and 141 clips, respectively. The gains come mainly from aesthetic quality, imaging quality, IAA, and IQA, and UniPercept favors our outputs for all three input generators.

\begin{table}[H]
    \centering
    \normalfont\fontsize{6.4pt}{7.8pt}\selectfont
    \setlength{\tabcolsep}{1.1pt}
    \renewcommand{\arraystretch}{1.20}
    \caption{\textbf{Comparison with the official MiniMax H3 2K API on \refbench.} Results are reported for each input generator and for all clips. Both methods receive the same $1024\!\times\!576$ inputs and produce $2560\!\times\!1440$ videos. SC, BC, MS, DD, AQ, and IQ denote subject consistency, background consistency, motion smoothness, dynamic degree, aesthetic quality, and imaging quality. Better results are in \textbf{bold}.}
    \label{tab:h3-extension-results}
    \begin{tabularx}{\linewidth}{@{}l<{\hspace{5pt}}l *{11}{>{\centering\arraybackslash}X}@{}}
        \toprule
        \multirow{2}{*}{\textbf{Generator}}
        & \multirow{2}{*}{\textbf{Refiner}}
        & \multicolumn{7}{c}{\textbf{VBench}$\uparrow$}
        & \multicolumn{4}{c}{\textbf{UniPercept}$\uparrow$} \\
        \cmidrule(lr{0.25em}){3-9}\cmidrule(l{0.25em}){10-13}
        & & \textbf{SC} & \textbf{BC} & \textbf{MS} & \textbf{DD} & \textbf{AQ} & \textbf{IQ} & \textbf{AVG}
        & \textbf{IAA} & \textbf{IQA} & \textbf{ISTA} & \textbf{AVG} \\
        \midrule
        WAN~2.1 & Official H3 2K API & 0.93412 & \textbf{0.95742} & \textbf{0.98454} & 0.58000 & 0.59501 & 0.63109 & 0.78036 & 61.4569 & 63.1538 & 42.8219 & 55.8108 \\
        \rowcolor{nvgreenlight}
        & Ours & \textbf{0.93653} & 0.95653 & 0.98211 & \textbf{0.60000} & \textbf{0.62475} & \textbf{0.66337} & \textbf{0.79388} & \textbf{67.7819} & \textbf{68.0763} & \textbf{44.4206} & \textbf{60.0929} \\
        \midrule
        SANA-Video & Official H3 2K API & 0.93693 & 0.95506 & \textbf{0.98241} & \textbf{0.91837} & 0.58800 & 0.65507 & \textbf{0.83931} & 61.4965 & 63.9841 & 41.3587 & 55.6131 \\
        \rowcolor{nvgreenlight}
        & Ours & \textbf{0.93844} & \textbf{0.95541} & 0.97617 & 0.81633 & \textbf{0.62901} & \textbf{0.70407} & 0.83657 & \textbf{70.8450} & \textbf{73.9745} & \textbf{45.0702} & \textbf{63.2966} \\
        \midrule
        LTX-2 & Official H3 2K API & 0.88238 & \textbf{0.93829} & \textbf{0.98588} & 0.80000 & 0.52288 & 0.60046 & 0.78831 & 54.3963 & 51.5903 & 40.7582 & 48.9149 \\
        \rowcolor{nvgreenlight}
        & Ours & \textbf{0.89600} & 0.93818 & 0.98472 & \textbf{0.84000} & \textbf{0.56008} & \textbf{0.65973} & \textbf{0.81312} & \textbf{63.3250} & \textbf{61.3944} & \textbf{45.0452} & \textbf{56.5882} \\
        \midrule
        All & Official H3 2K API & 0.91768 & \textbf{0.95022} & \textbf{0.98429} & \textbf{0.76510} & 0.56850 & 0.62870 & 0.80242 & 59.1006 & 59.5465 & 41.6482 & 53.4317 \\
        \rowcolor{nvgreen}
        & \textbf{Ours} & \textbf{0.92356} & 0.95001 & 0.98103 & 0.75168 & \textbf{0.60445} & \textbf{0.67553} & \textbf{0.81438} & \textbf{67.2936} & \textbf{67.7737} & \textbf{44.8438} & \textbf{59.9704} \\
        \bottomrule
    \end{tabularx}
\end{table}

\begin{figure}[H]
    \centering
    \includegraphics[width=0.78\linewidth]{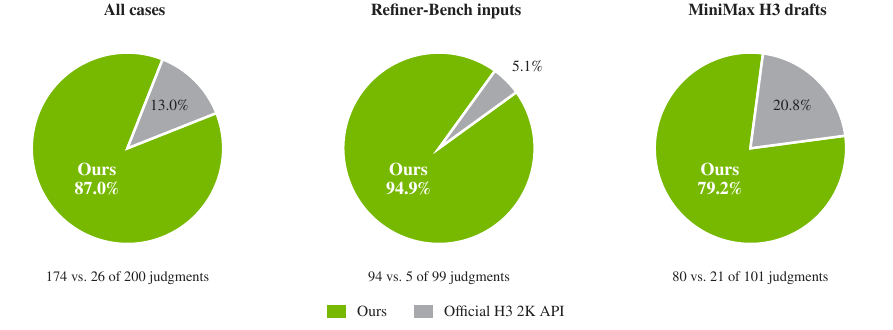}
    \caption{\textbf{User study against the official MiniMax H3 2K API.} Twenty participants each judge 10 cases, giving 200 judgments.}
    \label{fig:h3-user-study}
    \vspace{-4mm}
\end{figure}

\paragraph{User Study.} We conduct a blinded A/B user study on 50 cases: 25 \refbench{} clips and 25 MiniMax H3 drafts whose prompts come from X and Reddit. For each case, participants watch the input as a reference and choose the better of our output and the official API output, shown in random order. Twenty participants each judge 10 cases, giving 200 judgments. As shown in \cref{fig:h3-user-study}, participants prefer our outputs in 87.0\% of the judgments: 94.9\% for \refbench{} inputs and 79.2\% for MiniMax H3 drafts.

\paragraph{Latency.} \Cref{tab:h3-extension-latency} compares latency for a 5-second $2560\!\times\!1440$ clip. On one B200 GPU, latent upscaling, three refinement steps, and VAE decoding take 87.26 seconds without acceleration. Sol-Engine reduces the three refinement steps from 71.27 to 19.75 seconds through sparse attention, MXFP8 linear layers, and CUDA graphs, with sparse attention giving the largest saving, and its fast VAE decoder reduces decoding from 14.99 to 1.58 seconds. The full pipeline thus takes 22.45 seconds, $5.0\times$ faster than the official API at 112.7 seconds.

\begin{table}[H]
    \centering
    \normalfont\scriptsize
    \setlength{\tabcolsep}{4pt}
    \renewcommand{\arraystretch}{1.18}
    \caption{\textbf{Latency on a 5-second $2560\!\times\!1440$ clip.} Speedup is relative to the official API.}
    \label{tab:h3-extension-latency}
    \begin{tabularx}{\linewidth}{@{}l *{3}{>{\centering\arraybackslash}X}@{}}
        \toprule
        \textbf{Method} & \textbf{Hardware} & \textbf{Latency (s)}$\downarrow$ & \textbf{Speedup}$\uparrow$ \\
        \midrule
        Official H3 2K API & Undisclosed & 112.7 & 1.0$\times$ \\
        Ours & 1$\times$ B200 & 87.26 & 1.3$\times$ \\
        \rowcolor{nvgreen}
        \textbf{Ours + Sol-Engine} & 1$\times$ B200 & \textbf{22.45} & \textbf{5.0}$\times$ \\
        \bottomrule
    \end{tabularx}
\end{table}

\begin{figure}[H]
    \centering
    \begin{subfigure}[t]{\linewidth}
        \centering
        \includegraphics[width=\linewidth]{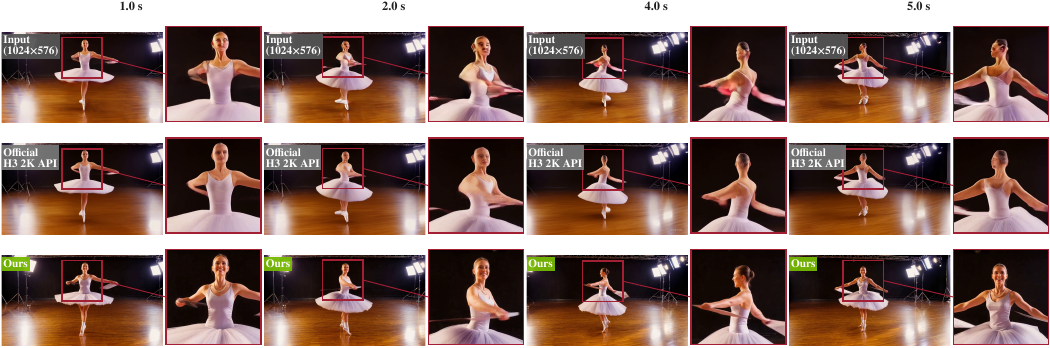}
        \caption{\textbf{\refbench{} input.}}
        \label{fig:h3-qualitative-ballet}
    \end{subfigure}
    \vspace{0.4em}
    \begin{subfigure}[t]{\linewidth}
        \centering
        \includegraphics[width=\linewidth]{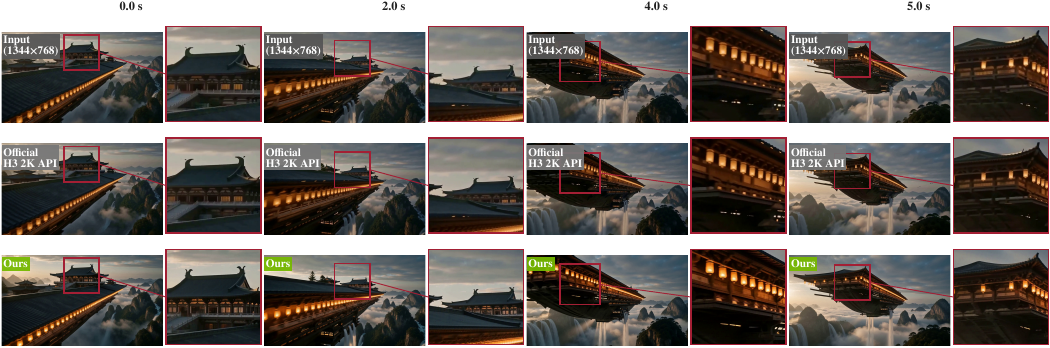}
        \caption{\textbf{H3 input.}}
        \label{fig:h3-qualitative-carrier}
    \end{subfigure}
    \caption{\textbf{Qualitative comparison with the official MiniMax H3 2K API.} Each panel shows the input, the official API output, and ours at four aligned frames, each followed by a magnified crop from the red box.}
    \label{fig:h3-qualitative}
\end{figure}

\paragraph{Qualitative Comparison.} \Cref{fig:h3-qualitative} shows two examples. In panel (a), a ballet clip from \refbench{}, the official API output loses the dancer's facial features and smears her arms during the turns, whereas our output keeps a clear face and clean arms. In panel (b), a MiniMax H3 draft of a flying palace carrier, the official output remains as soft as the draft, while ours renders sharper roof tiles, window lattices, and timber structure.

\FloatBarrier

\section{Limitations}
\label{sec:limitations}

On \refbench{}, the 23-step model achieves higher VBench and UniPercept averages than the one-step model. Our refinement objective targets local visual detail while preserving the base video's content and motion. Correcting large semantic, geometric, or motion errors is not an explicit training objective. Stage II evaluates three sampled frames per video using image-quality and preference reward models. These frame-based rewards do not directly measure long-range temporal consistency. \refbench{} covers AI-generated videos from several base generators; camera-captured degradations and broader video editing tasks remain outside this study.